\documentclass[a4paper,fleqn]{cas-dc}

\usepackage[numbers]{natbib}
\usepackage{graphicx}
\usepackage{amsmath,amssymb,amsfonts}
\usepackage{hyperref}
\usepackage[ruled,vlined]{algorithm2e}
\usepackage{multirow}
\usepackage{tabularx, booktabs}
\usepackage{subfig}
\usepackage{mathtools}
\usepackage{arydshln}
\usepackage{xcolor}
\usepackage{mwe}
\usepackage{floatrow}
\usepackage{xcolor}
\definecolor{red}{rgb}{1.00,0.00,0.00}
\definecolor{blue}{rgb}{0.00,0.00,1.00}

\newcommand{\cblue}[1] {\textcolor{blue}{#1}}

\def\tsc#1{\csdef{#1}{\textsc{\lowercase{#1}}\xspace}}
\tsc{WGM}
\tsc{QE}
\tsc{EP}
\tsc{PMS}
\tsc{BEC}
\tsc{DE}

\begin{document}

\let\WriteBookmarks\relax
\def\floatpagepagefraction{1}
\def\textpagefraction{.001}

\shorttitle{MVGrasp: Real-Time Multi-View 3D Object Grasping in Highly Cluttered Environments}
\shortauthors{H. Kasaei et~al.}

\title [mode = title]{MVGrasp: Real-Time Multi-View 3D Object Grasping in Highly Cluttered Environments}      

\tnotemark[1]

\author[1]{Hamidreza Kasaei}
\cormark[1]
\ead{hamidreza.kasaei@rug.nl}
\ead[url]{https://www.ai.rug.nl/irl-lab/}


\address[1]{Department of Artificial Intelligence, Bernoulli Institute, Faculty of Science and Engineering, University of Groningen, The Netherlands.}

\author[2]{Mohammadreza Kasaei}
\address[2]{School of Informatics, University of Edinburgh, UK}

\begin{abstract}
Nowadays robots play an increasingly important role in our daily life. In human-centered environments, robots often encounter piles of objects, packed items, or isolated objects. Therefore, a robot must be able to grasp and manipulate different objects in various situations to help humans with daily tasks. In this paper, we propose a multi-view deep learning approach to handle robust object grasping in human-centric domains. In particular, our approach takes a point cloud of an arbitrary object as an input, and then, generates orthographic views of the given object. The obtained views are finally used to estimate pixel-wise grasp synthesis for each object. We train the model end-to-end using a synthetic object grasp dataset and test it on both simulation and real-world data without any further fine-tuning. To evaluate the performance of the proposed approach, we performed extensive sets of experiments in four everyday scenarios, including isolated objects, packed items, pile of objects, and highly cluttered scenes. Experimental results show that our approach performed very well in all simulation and real-robot scenarios. More specifically, the proposed approach outperforms previous state-of-the-art approaches and achieves a success rate of $>90\%$ in all simulated and real scenarios, except for the pile of objects which is $82\%$. Additionally, our method demonstrated reliable closed-loop grasping of novel objects in a variety of scene configurations. The video of our experiments can be found here: 
\href{https://youtu.be/c-4lzjbF7fY}{\cblue{https://youtu.be/c-4lzjbF7fY}}

\end{abstract}


\maketitle

\section{Introduction}

\begin{figure}[!t]
\vspace{2mm}
    \includegraphics[width=\linewidth, trim = 0cm 0cm 0cm 0cm clip=true]{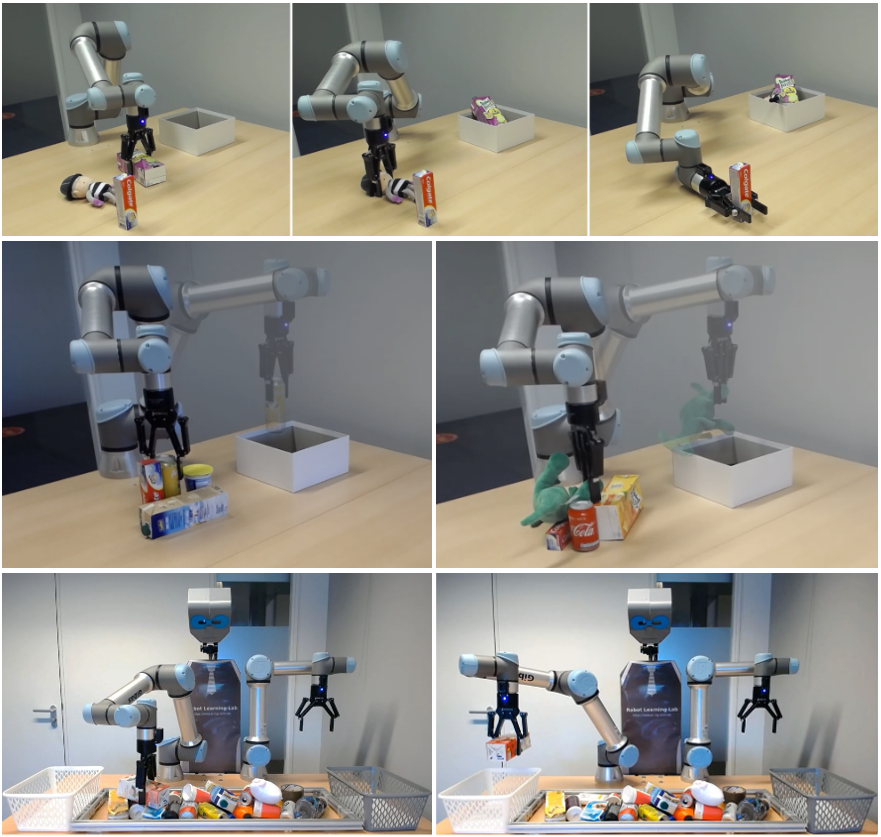}
    \caption{In human-centric environments, a robot often has to deal with four scenarios:  (\textit{top-row}) isolated cluttered objects, (\textit{second-row}) packed items and piles of objects, (\textit{bottom-row}) heavy-cluttered scenario. The robot should be able to predict a feasible grasp configuration for the target object based on the target object's pose and other objects in the scene. }
    \label{scenarios}
\end{figure}

Industrial robots are mainly designed to perform repetitive tasks in controlled environments. In recent years, there has been increasing interest in the deployment of service robots in human-centric environments~\cite{wirtz2021service,kasaei2021state,kasaei2018perceiving}. In such unstructured environments, object grasping is a challenging task due to the high demand for grasping a vast number of objects with a wide variety of shapes and sizes under various clutter and occlusion conditions (see Fig.~\ref{scenarios}). It is also expected that the robot could accomplish a given task as quickly as possible.

\begin{figure*}[!t]
    \centering
    \includegraphics[width=\textwidth]{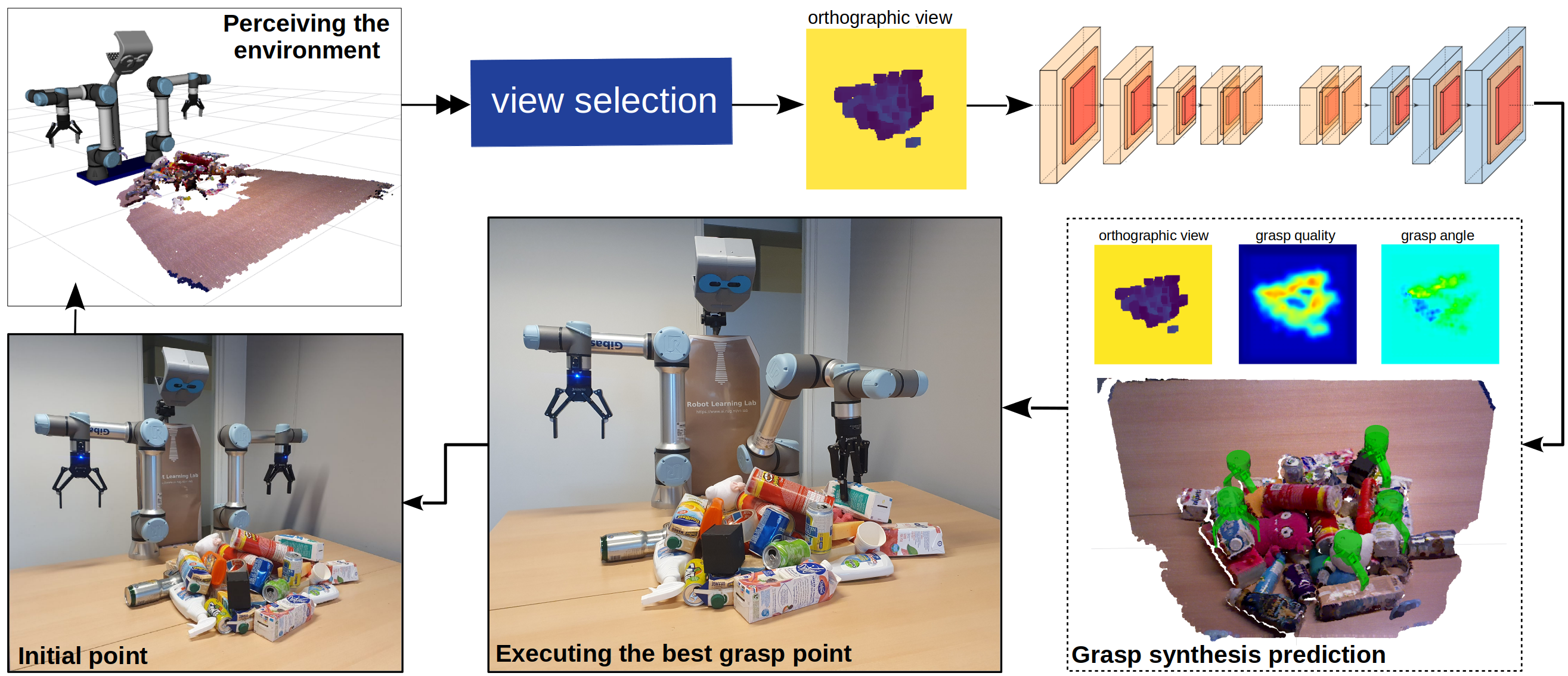}
  \caption{Overview of the proposed approach: we designed a mixed autoencoder for multi-view object grasping to be used in various everyday scenarios (isolated, packed objects, pile of objects). First, multiple views of a given scene are generated. The view selection module selects the best view in terms of grasping and then fed it into the grasp network to obtain a pixel-wise grasp synthesis. As shown in the most right part of the figure, the best grasp configurations are finally ranked and transformed from 2D to 3D using a set of known transformations. The robot then executes the best grasp configuration. This procedure is repeated until all objects get removed from the table.}
  \label{system_overview}
\end{figure*}

Although several grasping approaches have been developed successfully, many challenges remain. 
Recent works in grasp synthesis have mainly focused on developing end-to-end deep convolutional learning approaches to plan grasp configuration directly from sensor data. Although it has been proven that these approaches can outperform hand-crafted grasping methods, the grasp planning is mainly constrained to top-down grasps from a single depth sensor~\cite{mahler2017dex,morrison2018closing,morrison2020learning,kumra2020antipodal}. These approaches assume a favorable global camera placement and force the robot to grasp objects from a single direction, which is perpendicular to the image plane (4D grasp). Such constraints bound the flexibility of the robot, and the robot will not be able to grasp a range of household objects, e.g., bottles, boxes, etc. Furthermore, a certain set of objects has convenient parts to be grasped, e.g., the handle of a mug, or in some situation, it is easier to approach an object from different directions. Some of the deep-learning-based approaches take a very long time to sample and rank grasp candidates (e.g.,\cite{mahler2017dex}\cite{mousavian20196}), while others need to first explore the environment to acquire a full model of the scene and then generate point-wise 6D grasp configuration (e.g., Volumetric Grasping Network (VGN)~\cite{kumra2020antipodal}. Such 6D grasping approaches are mainly used in open-loop control scenarios and are not suitable for closed-loop scenarios.

In this work, we propose a real-time multi-view deep learning approach to handle object grasping in cluttered environments.  Our approach takes as an input a partial point cloud and generates multi-view depth images for each of the objects present in the scene. The obtained views are then passed to a view selection function. The best view is then fed to a deep network to estimate a pixel-wise grasp synthesis. Figure~\ref{system_overview} depicts an overview of our work. To summarize, the key contributions of this work are threefold:

\begin{itemize}
    \item We propose a novel deep learning architecture that receives a depth image as input and produces pixel-wise antipodal grasp synthesis as output for each object individually. We train the model end-to-end using a synthetic grasp dataset, and test it on both simulations and real-world data without any further fine-tuning.

    \item We show that the proposed approach is able to achieve reliable closed-loop grasping of novel objects across various scenes and domains.  
    Our approach, on average, could estimate stable grasp configurations in less than $25$ms. Therefore, the proposed approach is suitable to be used in real-time robotic applications that need closed-loop grasp planning. 
    
    \item We perform extensive sets of experiments in both simulated and real-robot settings to validate the performance of the proposed approach. In particular, we evaluate the proposed method on four common everyday situations, isolated objects, packed items, pile of objects, and highly cluttered scenarios. We demonstrate that our method outperforms previous state-of-the-art approaches and achieves a success rate of $>~90\%$ in all simulated and real scenarios, except for the pile of objects which is $82\%$.
     
\end{itemize}

\section{Related work}

Traditional object grasping approaches explicitly model how to grasp different objects mainly by considering prior knowledge about object shape, and pose~\cite{bohg2013data}. It has been proven that it is hard to obtain such prior information for never-seen-before objects in human-centric environments~\cite{kalashnikov2018qt,kasaei2019interactive}. Recent approaches address this limitation by formulating object grasping as an \textit{object-agnostic} problem, in which grasp synthesis is detected by visual features without taking prior \textit{object-specific} information into account. Therefore, these approaches are able to generalize the learned grasping features to novel objects. In this vein, much attention has been given to object grasping approaches based on Convolutional Neural networks (CNNs) ~\cite{lenz2015deep,mahler2017dex,morrison2018closing,breyer2020volumetric}. Deep-learning approaches for object grasping {falls} into two main categories depending on the input to the network: { \textit{Point-based} and \textit{View-based methods}}. In this section, we briefly review recent approaches to each category and refer the reader to a comprehensive review paper on this topic~\cite{newbury2022deep}.

\noindent
\textbf{Point-based approaches}: In this category, objects are represented as either 3D voxel grid or point cloud data and then fed into to a CNN with 3D filter banks~\cite{lundell2020beyond,breyer2020volumetric,li2020learning,mousavian20196,varley2017shape}. Some approaches first estimate the complete shape of the target object using a variational autoencoder network (e.g., \cite{lundell2020beyond, lundell2019robust,watkins2019multi,varley2017shape}). In other approaches, the robot first moves to various positions to capture different views of the scene, and then the obtained views are combined to create a complete 3D model of the scene (e.g., \cite{breyer2020volumetric, jiang2021synergies}). There are other methods that use machine learning approach to predict the grasp configuration from a partial view of the scene (e.g., \cite{mousavian20196,gualtieri2016high,li2020learning}). Unlike our approach, these approaches are often computationally expensive and not suitable for real-time closed-loop robotic applications. Furthermore, training such networks required enormous amount of data. 

Other approaches in this category use point cloud data directly~\cite{liang2019pointnetgpd,mousavian20196,gualtieri2016high}. One of the biggest bottlenecks with these approaches is the execution time and sensitivity to point cloud resolution. Unlike these methods, our approach generates virtual depth images of the object, and then generates grasp synthesis for the obtained object's views.

\noindent\textbf{View-based methods:} As an input to the network, some approaches use depth images. For instance, DexNet~\cite{mahler2017dex} and QT-Opt~\cite{kalashnikov2018qt} learn only top-down grasping based on depth images from a fixed static camera. Morrison et al.~\cite{morrison2018closing} proposed the Generative Grasping CNN (GG-CNN), a small neural network, which generates pixel-wise grasp configurations for a given depth image. Kumra et. al.,~\cite{kumra2020antipodal} developed GR-ConvNet, a \textit{large} deep network that generates pixel-wise grasp configurations for input RGB and depth data. While GR-ConvNet performed well on public grasp datasets, changing illumination and brightness affect its performance. Our approach predicts grasp configurations for each object, whereas the reviewed approaches generate grasp maps per scene. As we compute the grasp configuration in the object reference frame, the grasp prediction is independent of the camera pose while all reviewed approaches are highly dependent on the camera viewpoint (top-down view). Additionally, all the reviewed view-based approaches only work for top-down camera settings and are mainly focused on solving 4DoF grasping, which forces the gripper to approach objects from above. The major drawbacks of these approaches are inevitably restricted ways to interact with objects. Moreover, the robot is not able to immediately generalize to different task configurations without extensive retraining. We tackle these problems by proposing a multi-view approach for object grasping in highly crowded scenarios. We show that our model can be trained on small grasp dataset in an end-to-end manner and performed on both simulations and real-world data without any further fine-tuning.

\section{Problem Formulation}

We formulate grasp synthesis as a learning problem of planning parallel-jaw grasps for objects in clutter. In particular, we intend to learn a function that receives a collection of \textit{virtual} depth images of a 3D object as input, and returns as outputs the best view to approach the object, and a grasp map, which represents pixel-wise grasp configuration for the selected view. 
\subsection {Generating multiple views of objects}

A three-dimensional (3D) object is usually represented as point cloud, $p_i : i \in \{1,\dots,n\}$, where each point is described by their 3D coordinates $[x, y, z]$. To capture depth images from a 3D object, we need to set up a set of \textit{virtual} cameras around the object, where the $\textbf{Z}$ axes of cameras are pointing towards the centroid of the object. We first calculate the geometric center of the object as the average of all points of the object. Then, a local reference frame is created by applying principal component analysis on the normalized covariance matrix, $\Sigma$, of the projection of the object on the table, i.e., $\Sigma\textbf{V}=\textbf{EV}$, where $\textbf{E} = diag(e_1, e_2, e_3)$ contains the descending sorted eigenvalues, and $\textbf{V} = (\vec{v}_1, \vec{v}_2, \vec{v}_3)$ shows the eigenvectors. Therefore, $\vec{v}_1$, shows the largest variance of the points of the object. In this work,  $\vec{v}_1$ and the negative of the gravity vector are considered as $\textbf{X}$ and $\textbf{Z}$ axes, respectively. We define the $\textbf{Y}$ axis as the cross product of $\textbf{X} \times \textbf{Z}$. The object is then transformed to be placed in the reference frame. 

\begin{figure}[!t]
  \centering
    \includegraphics[width=\columnwidth, trim = 0cm 0cm 0cm 0cm clip=true]{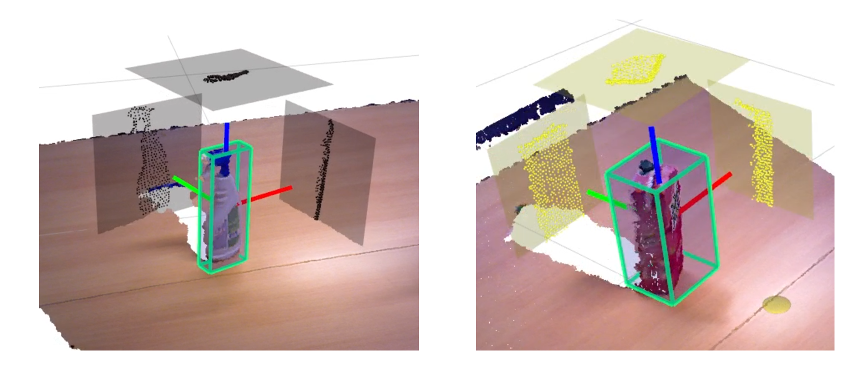}
    \caption{Two examples of generating bounding box, local reference frame, and three projected views for: (\textit{left}) a {glass-cleaner}; (\textit{right}) a {juice-box}. }
  \label{projections}
\end{figure}
From each virtual camera pose, we map the point cloud of the object into a  virtual  depth  image based on z-buffering. In particular, we first project the object to a square plane, $M$, centered on the camera’s center. The projection area is then divided into $l \times l$ square grid, where each bin is considered as a pixel. Finally, the minimum $z$ of all points falling into a bin is considered as the pixel value. In the case of object-agnostic grasping, since the grasp configurations depend on the pose and size of the target object, a view of the object should not be scale-invariant, we consider a fixed size projection plane ($l \times l$). In our setup, the size of each bin is defined as $f_g \times f_g$, where $f_g$ represents the size of the head of the robot's finger, and the length of each side of the projection plane is defined as $l \times f_g$. Figure~\ref{projections} shows the process of constructing local reference frames, generating bounding box and orthographic projections for two sample objects. 
 {It should be noted that the robot uses a single real Kinect camera to perceive the environment. Although our approach can work with arbitrary camera poses, the optimal camera pose is the one that offers greater scene coverage, as coverage directly affects the quality of virtual views. In particular, better scene coverage will make the object's point cloud more apparent, allowing us to generate more accurate view of the object. In our experiments, we realized that the ideal camera angle would be between 30 and 60 degrees with respect to the workplace, depending on the height of the camera.}

\subsection{View selection for grasping}
\begin{figure}[!t]
    \centering
    \begin{tabular}{ccc}
        \includegraphics[width=0.55\linewidth]{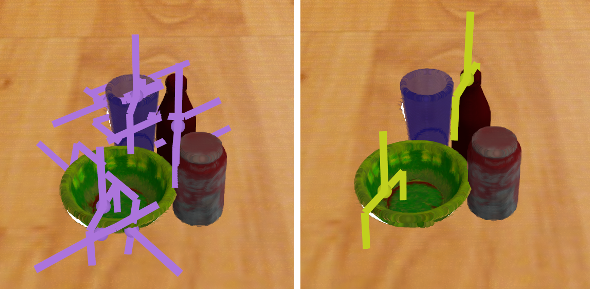}& \hspace{-5mm}
        \includegraphics[width=0.41\linewidth]{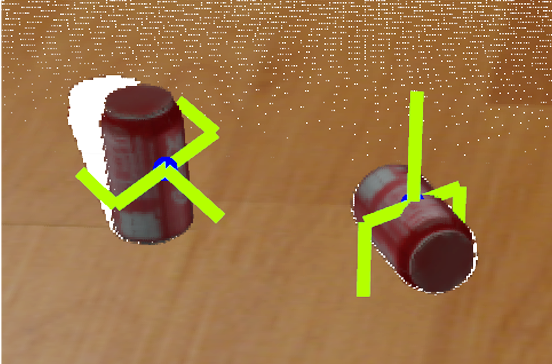}  
    \end{tabular}
  \caption{Examples of grasping objects in different situations: (\textit{left}) predicted grasp configurations for the given scene; (\textit{center}) grasps that are both kinematically feasible and collision free; (\textit{right}) the best grasp configuration for a coke-can object in two different situations.}
\label{different_grasps}
\end{figure}

View selection is crucial to make a multi-view approach computationally efficient. Although it is possible to pass all the views of the object into the network and then execute the grasp with a maximum score (Fig.~\ref{different_grasps} \textit{left}), such approaches are computationally expensive. In contrast, choosing a view that covers more of the target object's surface will not only reduce the computation time but also increase the likelihood of grasping the object successfully. Information theory provides a range of metrics (variance, entropy, etc.) from which the expected information gain can be calculated. Among these metrics, viewpoint entropy is a good proxy for expected information gain~\cite{thrun2002probabilistic}. In particular, viewpoints that observe the area of high entropy are likely to be more informative than those that observe low entropy areas. Therefore, we formulate our view ranking procedure based on viewpoint entropy. It  covers both the number of occupied pixels and the pixels' values. In particular, we calculate the entropy of a normalized projection view, $v$, by 
\begin{equation}
    H (v) = - \sum_{k=1}^{k^2} p_k \log_2(p_k)
\end{equation} 
\noindent where $p_k$ is the normalized value of pixel $k$, and $\sum_k p_k = 1$. The view with highest entropy is considered as the best view for grasping and then fed to the network to predict pixel-wise grasp configuration. We also consider the kinematic feasibility and distance that the robot would require to travel in the configuration space. In the case of large objects, or a pile of objects, there is a clear advantage (e.g., collision-free) to grasp from above (see Fig.~\ref{different_grasps} \textit{middle}), while for an isolated object, the direction of approaching object completely depends on the pose of the object relative to the camera (see Fig.~\ref{different_grasps} \textit{right}). The gripper approaches the object from an orthogonal direction to the projected view. It should be noted that the view selection function can be easily adapted to any other tasks' criteria. 

\vspace{-2mm}
\subsection {Network architecture}

We aim to learn a function that maps an input object's view to multiple outputs representing pixel-wise antipodal grasp configurations, $f_\theta: \mathcal{X} \rightarrow \mathbf{\mathcal{Y}}$. Towards this goal, we have designed a convolutional autoencoder that receives a depth image with height $h$ and width $w$ pixels as input, $x_{(i,j)} \in \mathbb{R}^{h \times w}$, and returns a pixel-wise grasp configuration map, $\textbf{G}$, i.e., $y_{(i,j)} = [\textbf{G}_{(i,j)}]$. 
The network is parameterized by its weights $\theta$. Our model is a single-input multiple-outputs network and is constructed using three types of layers, including {convolution}, {deconvolution}, and {batch normalization}. 

The encoder part is composed of six convolutional layers (C$1$ to C$6$). We use a Rectified Linear Unit ({ReLU}) as the activation function of all layers of the encoder part to force the negative values to zero and eliminating the vanishing gradient problem which is observed in other types of activation functions. We added a batch normalization layer after each convolutional layer to stabilize the learning process and reducing the number of training epochs by keeping the mean output and standard deviation close to $0$ and $1$, respectively. The decoder part is composed of six transposed convolutional layers (T$1$ to T$6$), followed by three separate output layers for predicting grasp quality, width, and rotation. Similar to the encoder part, we use the ReLU activation function for all layers and add a batch normalization after each layer. We use the same padding in all convolution and transposed convolution layers to make the input and output be of the same size (see Section~\ref{exprimental_result}). 

\subsection{Grasp execution}

In this work,  an antipodal grasp point is represented as a tuple, $g_i = \langle (u, v), \phi_i, w_i, q_i\rangle$, where $(u, v)$ stands the center of grasp in image frame, $\phi_i$ represents the rotation of the gripper around the Z axis in the range of $[\frac{-\pi}{2}, \frac{\pi}{2}]$, $w_i$ shows the width of the gripper where  $w_i \in [0, w_{max}]$, and the success probability of the grasp is represented by $q_i \in [0,1]$. Given an input view, the network generates multiple outputs, including: $(\boldmath{\phi}, \textbf{W}, \textbf{Q}) \in \mathbb{R}^{h \times w}$, where pixel value of images indicate the measure of $\phi_i, w_i, q_i$ respectively. Therefore, from $f_\theta (I_i) = \mathbf{G}_i$, the best grasp configuration, $\operatorname{g^*}$, is the one with maximum quality, and its coordinate indicate the center of grasp, i.e., $(u,v) \leftarrow \operatorname{g^*} = \operatorname*{argmax}_\mathbf{Q} ~ \mathbf{G}_i$. Given a grasp object dataset, $\mathcal{D}$,  containing $n_d$ images,  $D = \{(x_i, y_i) | 1 \le i \le n_d\}$, our model can be trained end-to-end to learn $f_\theta (.)$. 

After obtaining the grasp map of an input view, the Cartesian position of the selected grasp point, $(u,v)$, can be transformed  to object's reference frame since the transformation of the orthographic view relative to the object is known. The depth value of the grasp point is estimated based on the minimum depth value of the surrounding neighbors of $(u,v)$  that are within a radius of $\Delta$, where $\Delta = 5$mm.  Finally, the robot is instructed to perform grasping action.

\section{Results and Discussions}
\label{exprimental_result}

We performed several rounds of simulation and real-robot experiments to evaluate the performance of the proposed approach. We were pursuing three goals in these experiments: (\textit{i}) evaluating the performance of object grasping in four everyday scenarios, including: isolated object, packed objects, pile of objects, and highly cluttered scenes; (\textit{ii}) investigating the usefulness of formulating object grasping as an object-agnostic problem for general purpose tasks; (\textit{iii}) determine whether the same network can be used in both simulation and real-robot systems without additional fine-tuning. Towards this goal, we employed the same code, and used the same grasp prediction network in both simulation and real-robot experiments, in which the network was trained on a synthetic grasp dataset. 

\begin{figure}[!t]
    \includegraphics[width=\linewidth, trim = 0cm 0cm 0cm 0cm clip=true]{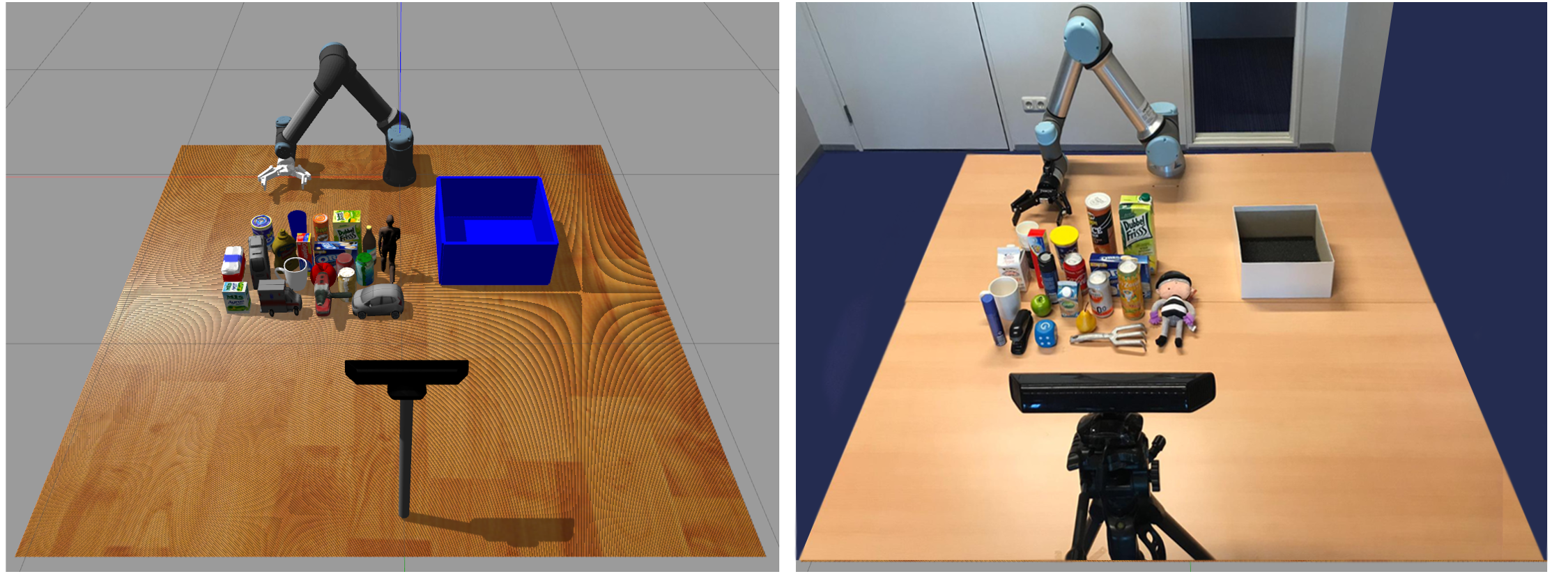}\vspace{1mm}\\
    \caption{Our experimental setups in: (\textit{left}) simulation environment; and (\textit{right}) the real-world setting. It should be noted, both simulated and real sets of objects used for evaluations are shown in these figures.}
  \label{exp_setup}
\end{figure}

\subsection{Experimental setup}
Figure \ref{exp_setup} shows our experimental setup in simulation (\textit{left}) and real-robot (\textit{right}). In this work, we have developed a simulation environment in Gazebo similar to our real-robot setup to extensively evaluate performance of our approach as well as its ability to adapt to different test environment settings. The robot and the camera in the simulated environment were placed according to the real-robot setup to obtain consistent performance. Our setup consists of a Universal Robot (UR5e) with a two-fingered Robotiq 2F-140 gripper, a Kinect camera mounted on a tripod, and a user interface to start and stop the experiments. 

To generate a multi-view grasp dataset and assess the performance of the proposed approach, we designed a clear table task, where the robot has to pick up all objects from its workspace and put them into a \textit{basket}. In all experiments, the robot knows in advance the pose of the  \textit{basket} as the placing position. We remove the table plane from the measured point cloud, then cluster the remaining points~\cite{rusu20113d,kasaei2018towards}. We consider each of the cluster as an object candidate. Note that object segmentation is beyond the scope of this paper and more advance technique can be considered, e.g., \cite{xiang2020learning}. 

At the beginning of each experiment, we set the robot to a pre-defined setting, and randomly place objects on the table. Afterward, the robot needs to predict grasp synthesis and select the best graspable pose of the target object, picks it up, and puts it in the \textit{basket}. This procedure is repeated until all objects get removed from the table, or the robot can not find an executable grasp point for the present objects. We use RRT motion planner to check for a collision-free path to each grasp pose and execute the one with the highest quality score. We assessed the performance of our approach by measuring success rate, i.e., $\frac{number~of~successful~grasps}{number~of~attempts}$. If the robot could place the object inside the basket we consider it as a success, otherwise a failure. If non of the predicted grasp points can be executed we consider the trial a failure too.  
For grasping experiments in simulation, we used $20$ simulated objects, imported from different resources (e.g., the YCB dataset~\cite{calli2017yale}, Gazebo repository, and etc). For real-robot experiments, we used $20$ daily-life objects with different materials, shapes, sizes, and weight (see Fig.~\ref{exp_setup}). All objects were inspected to be sure that at least one side fits within the gripper. All tests were performed with a PC running Ubuntu $18.04$ with a $3.20$ GHz Intel Xeon(R) $i7$, and a Quadro P$5000$ NVIDIA.

\begin{figure}[!b]
    \includegraphics[width=\linewidth, trim = 0cm 0cm 0cm 0cm clip=true]{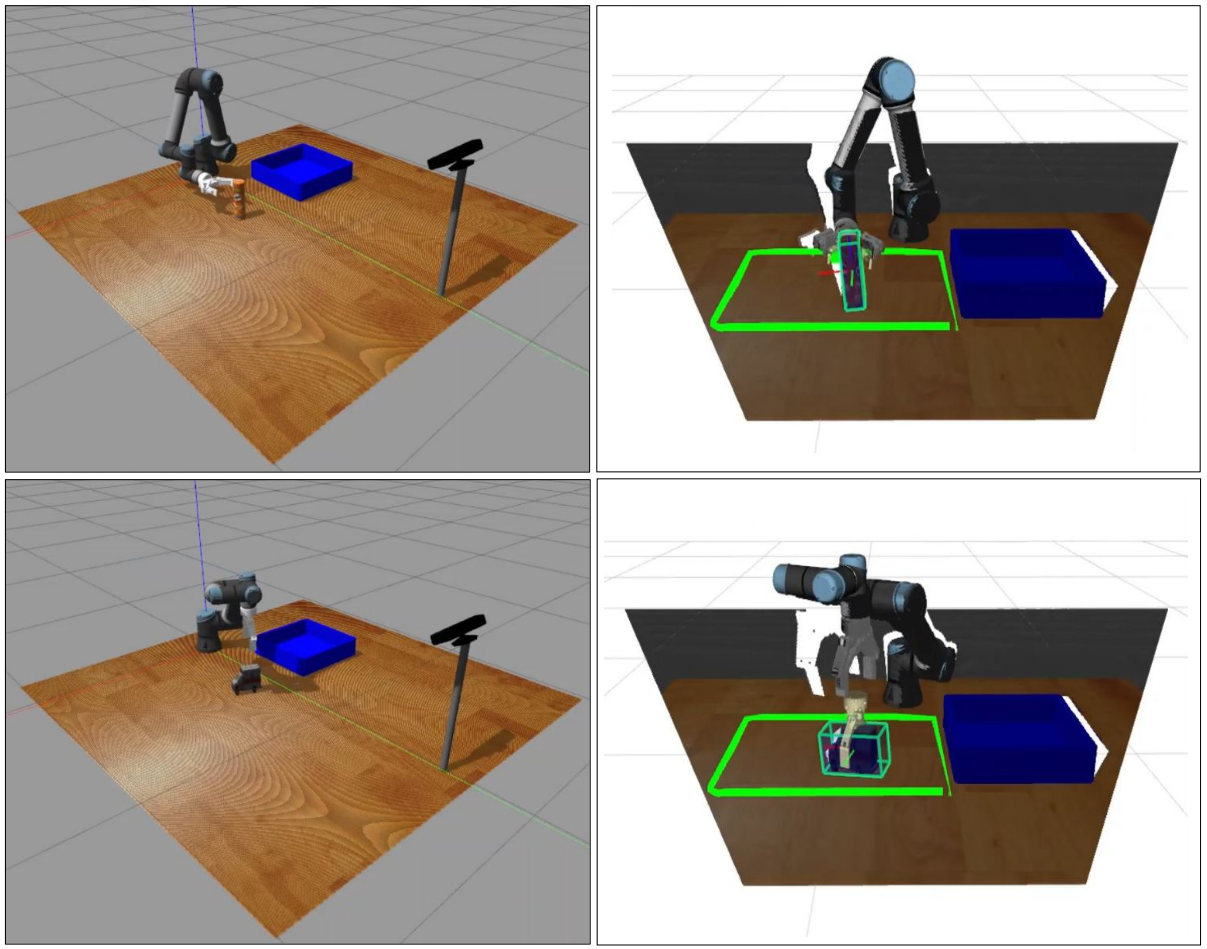}\\
    \caption{Generating synthetic grasp dataset: (\textit{left-column}) We randomly spawn an object in the workspace of the robot; then we instruct the robot to grasp the object and place it into the basket. (\textit{right-column}) The workspace of the robot is shown in green. The robot iteratively selects one of the extracted views of the object to approach and grasp the object. We record all positive samples and discard those configurations that lead to a collision with the object or the table (negative samples). }
  \label{dataset_setup}
\end{figure}

\subsection{Multi-view grasp dataset generation}

\begin{figure}[!t]
    \includegraphics[width=\linewidth, trim = 0cm 0cm 0cm 0cm clip=true]{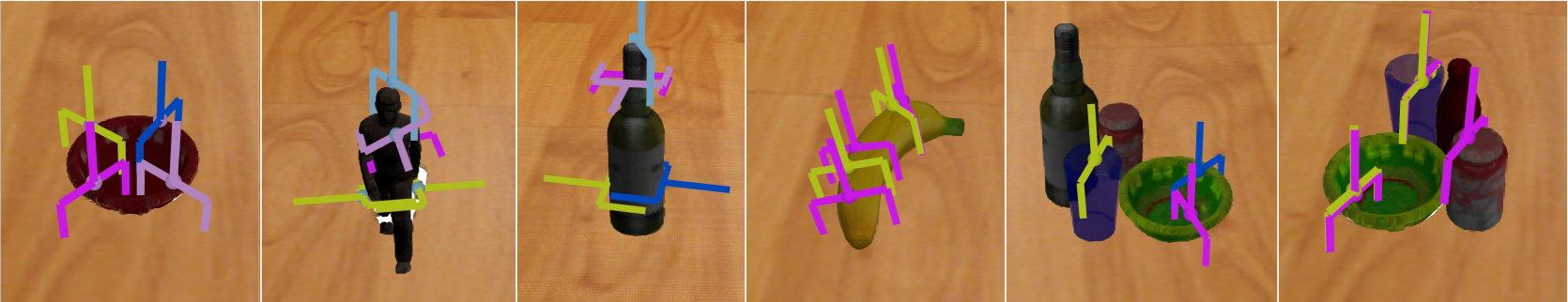}\\
    \caption{Examples of generated grasp synthesis for various objects in different scenarios.}
  \label{fig:grasp_example}
\end{figure}

In order to generate a synthetic dataset, we randomly spawn an object in the workspace of the robot as shown in Fig.~\ref{dataset_setup}. The robot then detects the object and extracts the multiple views of the object. In order to obtain a ground truth grasp configuration, we randomly sample different grasp configurations for each of the extracted view of the object. We then convert each grasp configuration to 3D space and instruct the robot to approach the object. Afterwards, we try to optimize the selected grasp configuration using simulated annealing~\cite{kirkpatrick1983optimization} by iteratively updating the orientation and width of the gripper and computing a fitness value based on three main factors: (\textit{i}) the proportion of object's points that are between the gripper's fingers relative to all object's points (coverage criteria); (\textit{ii}) how stable the point is, which is measured based on how well the normals of the fingers overlap with the normals of the selected points between the two fingers; and (\textit{iii}) we also considered the distance of the selected grasp point to the center of projected view. Furthermore, to make sure that the obtained grasp configuration is stable enough during manipulation phase, we instruct the robot to place the object into the blue basket. To extend the size of dataset and cover various objects with different shape and size, we formed packed and pile of objects scenes using four to six objects and generate grasp configurations for those scenes in addition to generating grasp synthesis for isolated object scenario. Using the described procedure, we generate a grasp dataset of approximately one million positive grasp configuration and discard those configurations that lead to a collision with the object or the table (negative samples). Examples of generated grasp synthesis for different objects are depicted in Fig.~\ref{fig:grasp_example}. 

\subsection{Ablation studies}
We study the impact of each of the key components our approach, i.e, network architecture and view selection strategy, through several ablation experiments. We compared our approach with five grasp prediction baselines, including DexNet~\cite{mahler2017dex}, GG-CNN~\cite{morrison2018closing}, GR-ConvNet~\cite{kumra2020antipodal}, Morrison et al.~\cite{morrison2020learning}, and Grasp Pose Detection (GPD)~\cite{gualtieri2016high}. 
In our experiments, DexNet~\cite{mahler2017dex}, GG-CNN~\cite{morrison2018closing}, GR-ConvNet~\cite{kumra2020antipodal}, and Morrison et al.~\cite{morrison2020learning} have access to global projected top-down view of the full scene, while our approach uses extracted views of the object. GPD uses the partial point cloud of the object as input.

\subsubsection{Network architecture}

We trained several networks with the proposed architecture but different parameters including filter size, dropout rate, loss functions, optimizer, and various learning rates for $200$ epochs each. The $80\%$ of the generated grasp dataset is used for training and the remaining $20\%$ is used as the evaluation set. Data augmentation techniques such as random photometric distortion, random clipping, and multi-scale resizing were also applied. We used Intersection over Union (IoU) metric. {In particular, a grasp pose is considered as a valid grasp if the intersection of the predicted grasp rectangle and the ground truth rectangle is more than $\alpha \%$, and the orientation difference between predicted and ground truth grasp rectangles is less than $\beta$ degrees. We performed a set of experiments to evaluate the effect of IoU and orientation threshold. We observed that by increasing the $\alpha$ value and decreasing the $\beta$ threshold, the performance of grasp prediction decreased as more match prediction with the ground truth will be considered as true positive. We experimentally found that IoU of $25\%$ and $30$ degrees threshold lead to the best results.} The final architecture is shaped as: C$_{(9 \times 9 \times 8,~S_3)}$, C$_{(5 \times 5 \times 16,~S_2)}$, C$_{(5 \times 5 \times 16,~S_2)}$, C$_{(3 \times 3 \times 32)}$, C$_{(3 \times 3 \times 32)}$, C$_{(3 \times 3 \times 32)}$, T$_{(3 \times 3 \times 32)}$, T$_{(3 \times 3 \times 32)}$, T$_{(3 \times 3 \times 32)}$, T$_{(5 \times 5 \times 16,~S_2)}$, ~ T$_{(5 \times 5 \times 32,~S_2)}$, \\T$_{(9 \times 9 \times 32,~S_3)}$, where $S$ stands for strides. We used Adam optimizer with a learning rate of $0.001$, and Mean Squared Error as a loss function. Overall, the network was able to achieved grasp accuracy of $93\%$ on our synthetic grasp dataset. {It should be noted that by selecting non-optimized parameters, the quality of grasp prediction will be decreased, which will have an effect on the grasp success rate consequently.}

\subsubsection{View selection for grasping}

Another key component of the proposed method is the view selection strategy for grasping purposes. Therefore, we performed extensive sets of experiments in the context of isolated object removal task to study the impact of view selection on grasping. For these experiments, we randomly placed an arbitrary object inside the robot's workspace, and instruct the robot to grasp and put the object into the \textit{basket}.

\begin{figure*}[!t]
    \vspace{-1mm}
    \includegraphics[width=\linewidth, trim = 0cm 0cm 0cm 0cm clip=true]{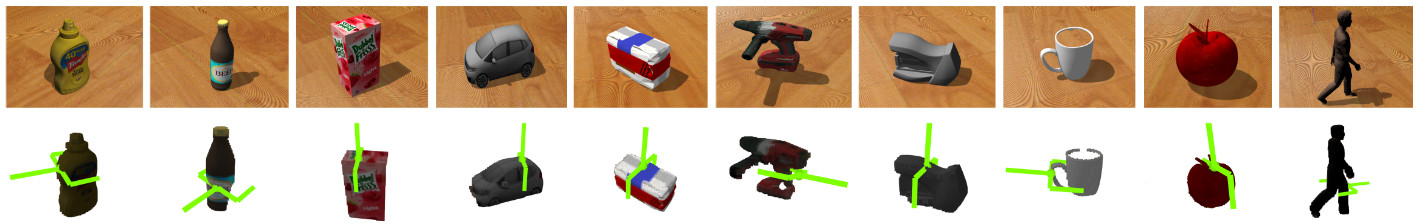}
    
    \caption{Qualitative results for grasping 10 household objects in the isolated scenario: (\textit{top-row}) visualizing objects in Gazebo and (\textit{lower-row}) their best grasp configurations. These results showed that our approach learned very-well the intended object-agnostic grasp function.}
    \label{isolated_exps}
\end{figure*}

\begin{table}[!t]
\newcolumntype{?}{!{\vrule width 0.5pt}}
\setlength\arrayrulewidth{0.5pt}
    \caption{Grasping performance in isolated object scenario.}
    \label{table:isolated_exps}
    \resizebox{0.9\linewidth}{!}{
    \begin{tabular}{|c|c|c|}
    \hline
    \textbf{Method} & \textbf{Type} & \textbf{Success rate ($\%$)}\\
    \hline\hline
    GPD & sim  & 78.7 (787/1000) \\\hline
    GGCNN & sim  & 72.6 (726/1000) \\\hline
    Morrison et al. & sim  & 77.1 (771/1000) \\\hline
    DexNet & sim & 79.4 (794/1000) \\\hline
    GR-ConvNet & sim & 81.4 (814/1000) \\\hline
    Ours~(top-down) & sim  & 73.2 (732/1000) \\\hline
    Ours~(random) & sim  & 51.3 (513/1000) \\\hline
    Ours & sim  & \textbf{92.6} (926/1000) \\\hline\hline
    GPD & real & 81.0 (81/100) \\\hline
    GGCNN & real & 78.0 (78/100) \\\hline
    Morrison et al. & real  & 77.0 (77/100) \\\hline
    DexNet & real & 81.0 (81/100) \\\hline
    GR-ConvNet & real & 84.0 (84/100) \\\hline
    Ours~(top-down) & real  & 67.0 (67/100) \\\hline
    Ours~(random) & real  & 49.0 (49/100) \\\hline
    Ours & real & \textbf{93.0} (93/100) \\\hline
    \end{tabular}}
\end{table}
Each object was tested $50$ times in simulation and $5$ times in a real environment. To speed up the real-robot experiments, we randomly placed five objects on the table. In each execution cycle, the robot selected the nearest object to the base and tried to remove it from the table. Results are reported in Table~\ref{table:isolated_exps}. By comparing both real and simulation experiments, it is visible that our approach outperformed all the selected approaches by a substantial margin. Particularly, in the case of simulation experiments, we achieved a grasp success rate of $92.6\%$ (i.e., $926$ success out of $1000$ trials), while GPD, GGCNN, DexNet, and GR-ConvNet obtained $78.7\%$, $72.6\%$, $79.4\%$ and $81.4\%$, respectively. The additional residual layers between the encoder and decoder parts of the GR-ConvNet architecture may help to explain why GR-ConvNet performs better than GGCNN, DexNet, and GPD. Furthermore, we realized that while our approach, GGCNN, and GR-ConvNet are suitable for closed-loop real-time scenarios ($> 45$ Hz, tested on the mentioned hardware), DexNet often took a long time to predict grasp candidates for a given input. It was due to the fact that the DexNet model contains a sampling routine that often takes a long time depending on the complexity of the scene (e.g., shape of the object and also number of objects in the scene). 
In the case of GGCNN, DexNet, and GR-CovNet, we observed that the majority of failures are caused by predicting the center of grasp point at the edge of an object. This can cause round objects to be forced out of the gripper as the gripper closes. It should be noted that even very small transformation errors can amplify such problems and lead to improper grasping of objects by the robot. In contrast, since our approach selects the best direction to grasp the object, such failures happened rarely. Fig.~\ref{isolated_exps} shows the best grasp configuration predicted by our approach for $10$ simulated objects.

In the case of real-robot experiments, the success rate of our approach was $93\%$ ($93$ success out of $100$ attempts), which was $12\%$, $15\%$, $16\%$, $11\%$, and $9\%$ better than GPD, GGCNN, Morrison et al., DexNet, GR-ConvNet, respectively. 
This was due to the fact that the proposed approach generated pixel-wise grasp configurations for the \textit{different views} of the target object, resulting in a variety of grasp options. This was not the case for the other approaches.  In particular, GPD often generated a few grasps for a target object and GGCNN, Morrison et al., DexNet, and GR-ConvNet returned \textit{only top-down} grasp configurations, in which, occasionally, none of the returned configurations led to a successful grasp. 

In the case of GGCNN, Morrison et al., DexNet, GR-ConvNet, and our approach with top-only view selection, failures mainly happened in grasping soda-can, bottle, human toy, and mustard object since the supporting area around the selected grasp point was too small and therefore, the object slipped and fall during manipulation. In the case of our approach with random view selection, the main failures were due to collision with the table, e.g., grasping a toppled soda-can from side. Some failures also occurred when one of the fingers of the gripper was tangent to the surface of the target object, which led to pushing the object away. In the case of our approach with view selection, the failed attempts were mainly due to inaccurate bounding box estimation, some objects in specific pose had a very low grasp quality score, and collision between the object and the bin (mainly happened for large objects e.g., \textit{Pringles} and \textit{Juice box}).  The experiments indicated that the proposed approach worked well for grasping isolated objects in both simulation and real-world environments without fine-tuning. In the following subsections, we test the performance of the approach in two challenging cluttered scenarios.


\subsection{Grasp evaluation on cluttered scenarios}
\label{grasp_results}

\begin{figure}[!b]
    \includegraphics[width=\linewidth, trim = 0cm 0cm 0cm 0cm clip=true]{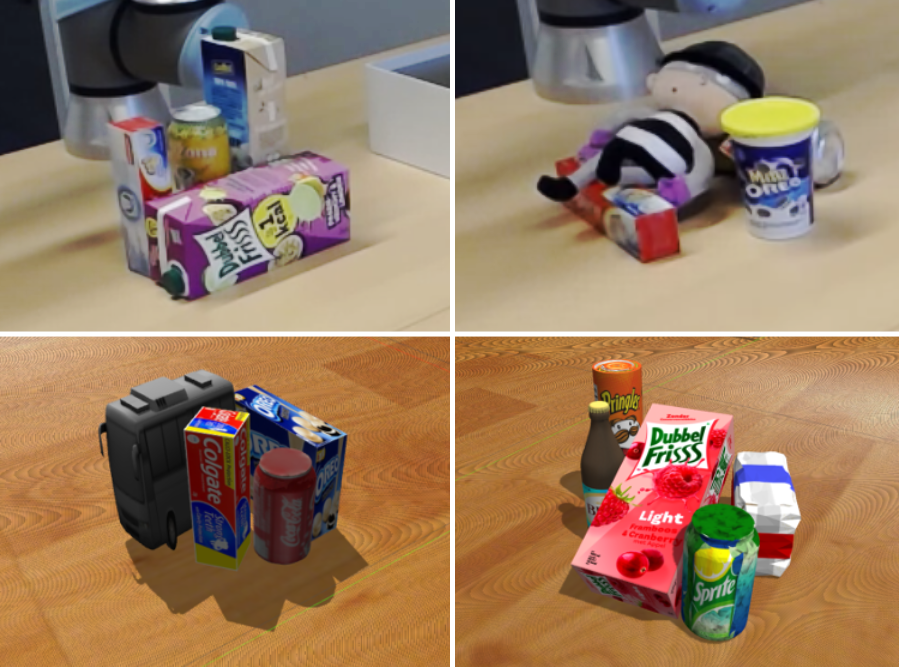}
    \caption{Illustrative examples of two cluttered scenarios in (top-row) real and (bottom-row) simulation  environments: (\textit{left-column}) packed objects; (\textit{right-column}) pile of objects.}
    \label{grasp_scenario}
\end{figure}

To assess the performance of the proposed approach in cluttered environments, we generate real and simulation scenes consisting of four to six objects for both packed and pile of objects scenarios. In the case of real-robot experiments, we put the selected objects into a box, then shake the box to remove bias, and finally pour the box in front of the robot to make a pile of objects. For packed of objects, we manually designed scenes by putting several objects tightly together (see Fig.~\ref{grasp_scenario} \textit{top-row}). To form experimental scenarios, we randomly selected instances of different objects.

In the case of simulation experiments, to generate scenes containing a pile of objects, we randomly spawn objects into a box placed on top of the table. We wait for a couple of seconds until all objects become stable, and then remove the box, resulting in a cluttered pile of objects. To construct a packed objects scenario, we iteratively placed a set of objects next together in the workspace. To form packed and pile of objects scenes, we randomly used $100$ instances of objects from our dataset. An example for each simulated scenario is shown in Fig.~\ref{grasp_scenario} \textit{bottom-row}). 

\begin{figure*}[!t]
    \includegraphics[width=\linewidth, trim = 0cm 0cm 0cm 0cm clip=true]{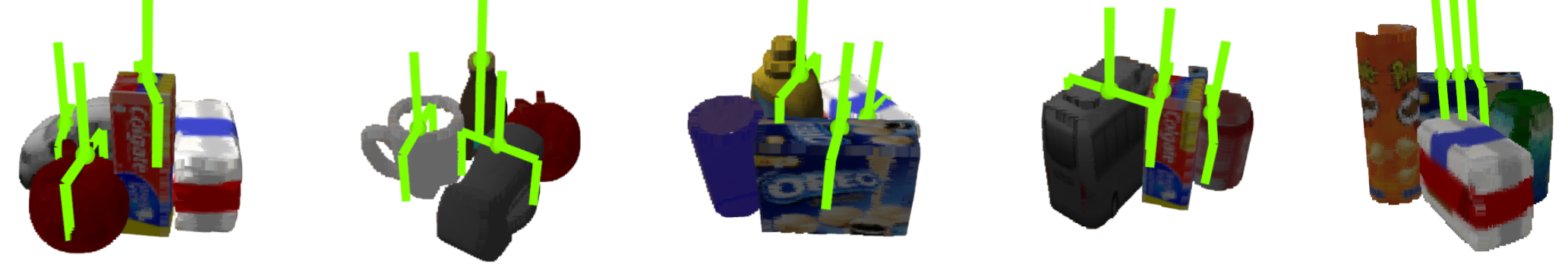}
    \caption{Qualitative results on packed scenarios: visualizing the top-three grasp configurations on five different densely-packed objects.}
    \label{packed_removing}
\end{figure*}

In this round of evaluation, in addition to the success rate, we report the average percentage of objects that are successfully removed from the workspace (average of number of objects removed from workspace at the end of a trial divided by initial number of objects). An experiment is continued until either all objects get removed from the workspace, or three failures occurred consecutively, or the quality of the grasp candidate is lower than a pre-defined threshed, $\tau \in \{ 0.8, 0.9\}$. In each execution cycle, the robot must execute the best possible grasp synthesis. Moreover, similar to the previous round of experiment, we performed similar experiments with GPD, GGCNN, Morrision et al, DexNet, and GR-ConvNet to compare their results with ours.

\begin{table}
\newcolumntype{?}{!{\vrule width 0.5pt}}
\setlength\arrayrulewidth{0.5pt}
    \caption{Grasping performance on packed objects scenarios.}
    \label{packed_exps}
    \resizebox{\linewidth}{!}{
    \begin{tabular}{|c|c|c|c|c|}
    \hline
    \textbf{Method} & \textbf{Type} & \textbf{Success rate} & \textbf{\#Attempts} & \textbf{Removal rate}\\
    \hline\hline
    GPD & sim & 0.54 & 128 & 0.69\\\hline
    GGCNN & sim & {0.55} & 131 & 0.72\\\hline
    Morrison et al. & sim & {0.61} & 122 & 0.74\\\hline
    DexNet & sim & {0.71} &114 & 0.81\\\hline
    GR-ConvNet & sim & {0.76} & 109 & 0.83\\\hline
    Ours ($\tau =0.8$) &  sim  & 0.84 & 105 & \textbf{0.88}\\\hline
    Ours ($\tau = 0.9$) & sim & \textbf{0.95} &75 & 0.71 \\\hline\hline
    
    GPD & real & 0.65 & 48 & 0.78 \\\hline 
    GGCNN & real & 0.46 & 54 & 0.63 \\\hline 
    Morrison et al. & real & {0.50} & 56 & 70.0 \\\hline 
    DexNet & real & {0.70} & 44 & 0.78 \\\hline 
    GR-ConvNet & real & {0.74} & 43 & 0.80 \\\hline 
    Ours ($\tau =0.8$) &  real  & \textbf{0.90}  & 42 & \textbf{0.95} \\\hline 
    \end{tabular}}
\end{table}
\subsubsection{Packed experiments}
Results are reported in Table~\ref{packed_exps}. Our approach outperformed all selected methods with a large margin in both simulated and real-robot experiments. It is clear from the results that by setting $\tau$ to $0.9$, the robot was able to remove fewer objects from the workspace while achieving a greater success rate, while setting $\tau$ to $0.8$ led to good balance between success rate and percentage of objects removed. Therefore, we use this configuration for real-robot experiments. The GPD struggles to find a grasp synthesis for small and flat objects. This causes the experiments to be discontinued early, thus negatively impacting the percentage of objects cleared. Most failure cases for GGCNN, Morrison et al., DexNet, and GR-ConvNet were either due to inaccurate grasp point prediction (e.g., edge of the object). This causes objects falling out of the gripper during manipulation, or the gripper running into other nearby objects. We also noticed some failures where the robot was unable to close the gripper or reach the target pose because other objects were present.

On closer inspection of real-robot experiments, we realized that the proposed method could successfully grasp $38$ objects out of $42$ attempts, resulting in $90\%$ success rate and $95\%$ percent cleared, while the second best approach (GR-ConvNet) achieved $74\%$ success rate and $80\%$ percent cleared. The RGB-D input modalities of GR-ConvNet could be another reason for achieving the second best performance. In contrast to GPD, GGCNN, DexNet, and Morrison et al., which only use depth information, GR-ConvNet also receives color information, which can help the network to get the notion of the object based on color information, while making the approach sensitive to the brightness of the environment. The GGCNN showed the worst results by executing $29$ unsuccessful grasp attempts, and achieving a $46$ percent success rate and $63$ percent clearance. DexNet, GPD, and Morrison et al. achieved the third, fourth, and fifth places, respectively. We found that mug-like objects and bottle-like objects are difficult to grasp for GPD and all top-down view based approaches, as the target object mostly slipped out of the gripper during the manipulation phase. We observed that the proposed approach is able to predict robust grasp quality scores for a given object. Figure~\ref{packed_removing} illustrates five examples of packed removal experiments.
\begin{figure}[!b]
    \includegraphics[width=\linewidth, trim = 0cm 0cm 0cm 0cm clip=true]{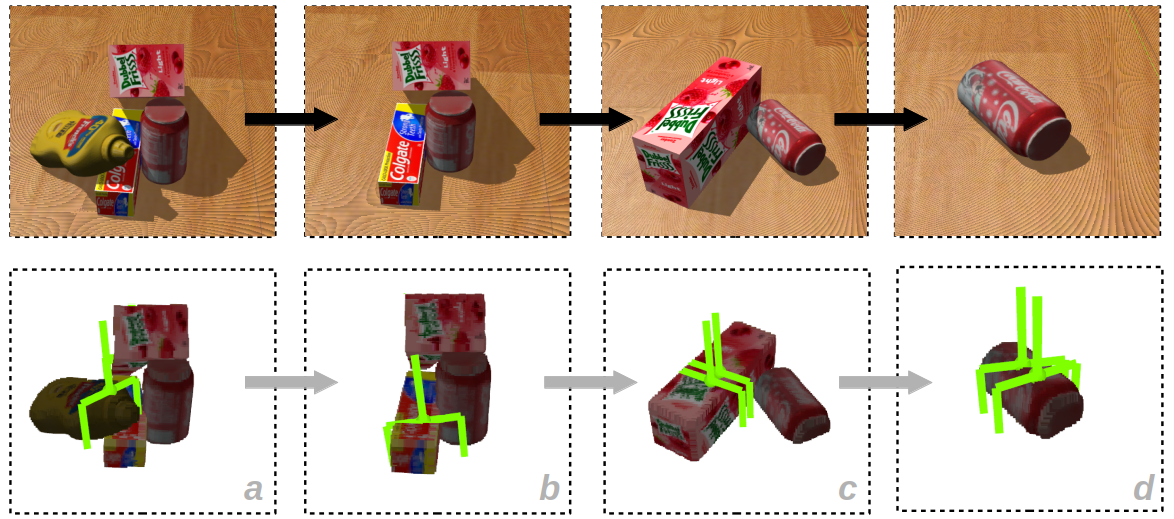}
    \caption{An example of successful sequence of removing a pile of objects: in this experiment, the robot successfully removed \textit{Mustard}, \textit{Colgate}, \textit{Juice box}, and \textit{Coke can} one by one. In each execution cycle, grasp configurations that are collision-free and kinematically feasible are shown by green color.} 
    \label{pile_removing}
\end{figure}

\subsubsection{Pile experiments} The obtained results are summarized in Table~\ref{pile_exps} and examples trial are shown in the accompanying video. We use an example to explain the results. Figure~\ref{pile_removing} depicts a successful sequence of removing a pile of four objects using the proposed approach. It was observed that after removing \textit{Mustard} and \textit{Colgate} objects from the workspace (Fig.~\ref{pile_removing} \textit{a, b}), the complexity of the scene reduced significantly. Therefore, the robot could find more grasp configurations whose grasp quality exceeds the threshold (Fig.~\ref{pile_removing} \textit{c, d}). As shown in this example, while the robot was interacting with an object, the pose of the other objects changed completely (e.g., toppled coke can and juice box), which make grasping the remaining object easier. 

\begin{table}[!t]
\newcolumntype{?}{!{\vrule width 0.5pt}}
\setlength\arrayrulewidth{0.5pt}
    \caption{Grasping performance on pile of objects scenarios}
    \label{pile_exps}
    \resizebox{\linewidth}{!}{
    \begin{tabular}{|c|c|c|c|c|}
    \hline
    \textbf{Method} & \textbf{Type} & \textbf{Success rate} & \textbf{\#Attempts} & \textbf{Removal rate}\\
    \hline\hline
        GPD & sim & 0.62 & 107 & 0.66 \\\hline 
    GGCNN & sim & 0.64 & 112 & 0.72 \\\hline 
    Morrison et al. & sim & 0.65 & 116 & 0.75 \\\hline 
    DexNet & sim & 0.72 & 111 & 0.80 \\\hline 
    GR-ConvNet & sim & 0.72 & 117 & 0.84 \\\hline 
    Ours ($\tau =0.8$) &  sim  & \textbf{0.75} & 119 & \textbf{0.89} \\\hline\hline 
    GPD & real & 0.55 & 51 & 0.70 \\\hline 
    GGCNN & real & 0.57 &54 & 0.78 \\\hline 
    {Morrison et al.} & real & 0.62 & 55 & 0.85 \\\hline 
    {DexNet} & real & 0.64 & 53 & 0.85 \\\hline 
    {GR-ConvNet} & real & 0.69 & 54 & \textbf{0.92} \\\hline 
    Ours ($\tau =0.8$) &  real  & \textbf{0.82} & 45 & \textbf{0.92} \\\hline 
    \end{tabular}}
\end{table}

\begin{figure}[!t]
    \includegraphics[width=\linewidth, trim = 0cm 0cm 0cm 0cm clip=true]{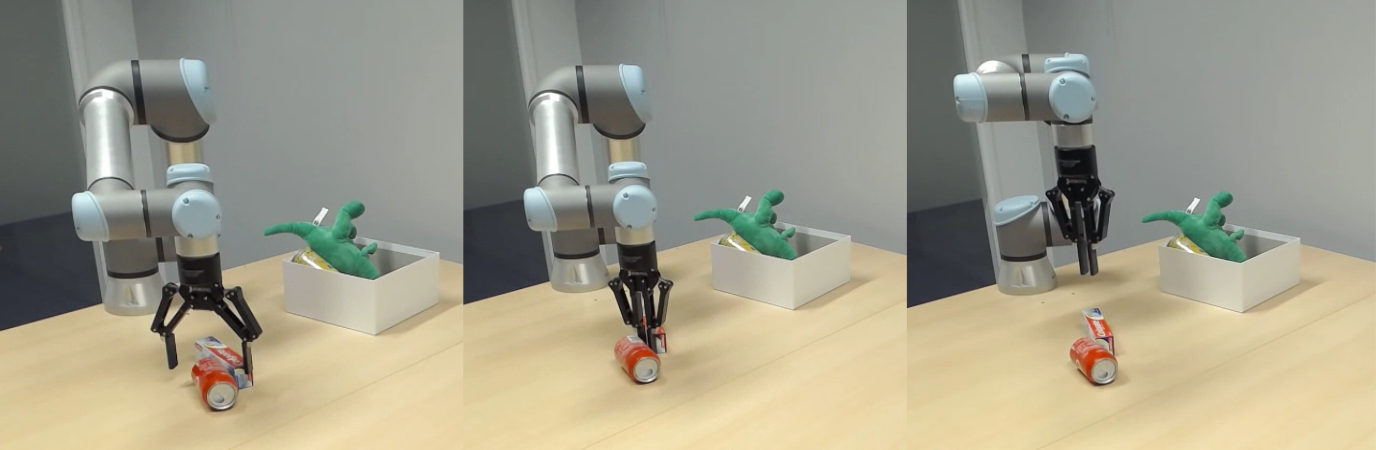}
    \caption{One of the reasons for grasp failures was detecting a small gap between two nearby objects as a graspable area. The execution of such grasp points often led to the objects being pushed apart. } 
    \label{failure_gap}
\end{figure}

In contrast, in some other cases, we observed that after grasping one of the object, the other objects moved into positions where they are no longer graspable or reachable, such as a toppled \textit{OreoBox} or an object that went out of the workspace. Such a situation was one of the main reasons for unsuccessful attempts. We also noticed that the robot occasionally tried to grasp multiple objects at once. By comparing all results it is visible that, in terms of success rate our approach works better than all the selected baselines in both simulation and real robot experiments. In simulation experiments, our approach had a $5\%$ removal rate advantage over GR-ConvNet and a $9\%$ advantage over DexNet; however, in real robot testing, GR-ConvNet performed just as well as our approach ($92\%$), and DexNet achieved an $85\%$ pile removal rate. In simulation experiments, DexNet and GR-ConvNet achieved $72\%$ success rate which was 3\% less than the success rate of our approach ($75\%$). The success rate for GPD, GGCNN, Morrision et al., in both simulation and real robot experiments, was less than $70\%$, as they predict false positive grasp points and unsuccessfully attempted those grasps three times in a row, ending the trials. This often happened for small objects as it was not always possible to infer more than one grasp synthesis for them. Some other failures occurred due to the lack of friction, applying limited force to the object, collision with other objects, and unstable grasps predictions. Other times, the robot mistook a tiny space between two objects as a graspable area, leading to failures. An example of such failures is depicted in Fig.~\ref{failure_gap}.


\subsection {Grasp evaluation in heavy cluttered scenarios}
\begin{figure*}[!t]
    \vspace{-3mm}
    \includegraphics[width=\linewidth, trim = 0cm 0cm 0cm 0cm clip=true]{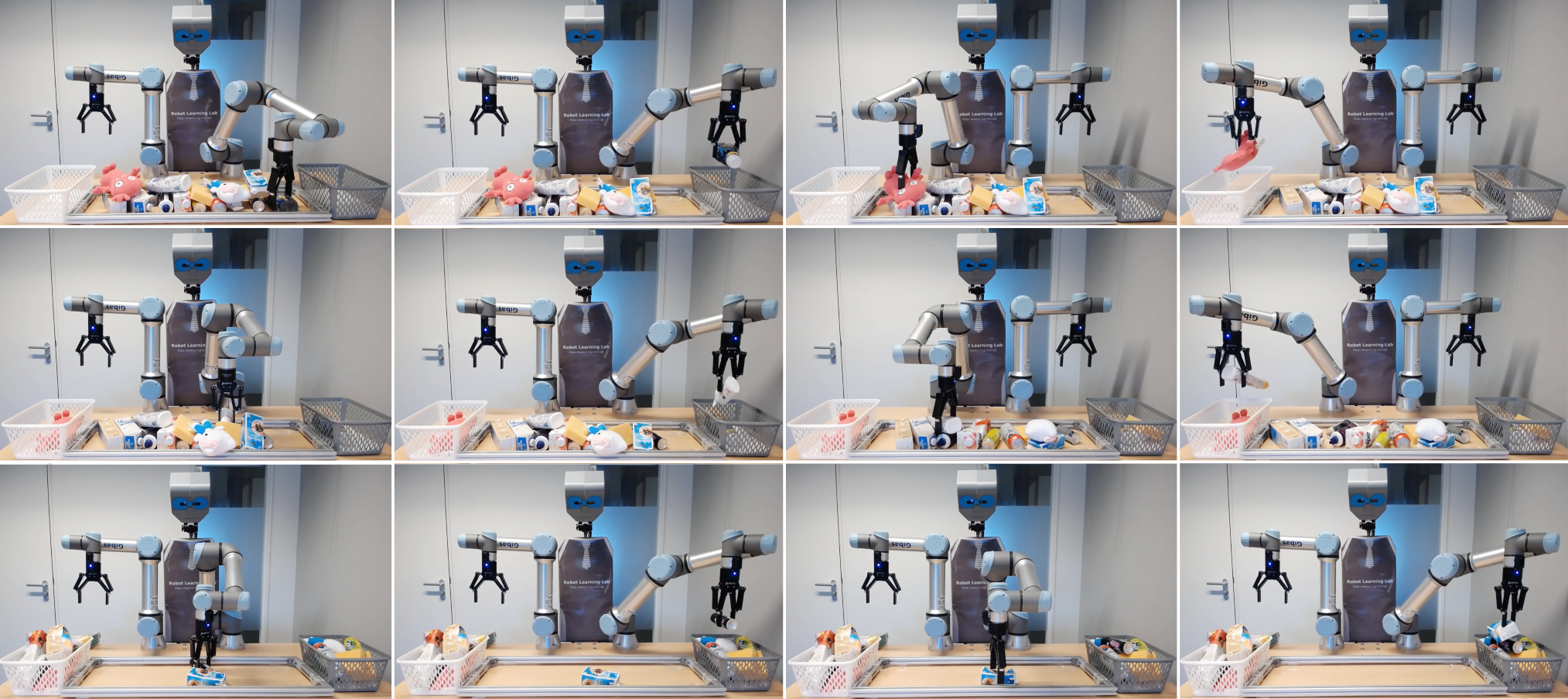}
    \caption{A sequence of snapshots showing how our robot could clean a heavy cluttered scene successfully ($> 15$ objects): In this experiment, the robot should remove all objects one by one from the table. In each execution cycle, the robot first detects all grasp synthesis for the scene and then executes the best one. The robot will place the object into the left basket if the selected grasp point is on the left side of the robot, otherwise it places the object into the right basket.  } 
    \label{highly_cluttered}
\end{figure*}

We also assess how well our method performs in heavy cluttered scenario, where we randomly place more than $15$ objects in the robot's workspace. Our experimental setups is shown in Fig.~\ref{highly_cluttered}. In this evaluation round, we use our dual arm robot which has two UR5e manipulators and an Asus-Xtion RGB-D camera. 
In contrast to the previous round of experiments where the camera was in front of the robot, the camera is in the robot side in this round of evaluation. It should be noted that in our approach, the grasp prediction is independent of the camera pose since we calculate the grasp synthesis in the object reference frame. In this experiment, the robot must take each item off the table one at a time and put it into one of the baskets. In each execution cycle, the robot first detects all grasp synthesis for the scene and then executes the best one. If the chosen grasp point is on the left side of the robot, the object will be placed in the left basket; if not, it will be placed in the right basket. We ran $10$ round of experiments and the robot could successfully put all of the objects into the baskets in 9 out of 10 experiments. Fig.~\ref{highly_cluttered} displays a series of snapshots of how our robot could clean a heavy cluttered scene successfully. These experiments showed that our approach can be easily adapted to any robotic platforms without any further training. Some of the grasp failures in this scenario can be attributed to the previously described issues, but the majority of them are either because of inaccurate grasp predictions, or the gripper colliding with surrounding objects or the environmental constraints imposed by the setup (aluminum frame and baskets). We observed that collisions with the baskets often happened due to the fact that the large objects were mostly occluded by the adjacent clutter pile and their stable graspable region (near center of the mass) were hidden. Therefore, the robot predict the grasp point at the edge of the object which causes the object to slip in the manipulation phase. Additionally, we observed that the gripper colliding with nearby objects could prohibit the robot from achieving the desired stance or from closing the gripper since the fingers would be obstructed. {By considering high-level information about the environment (e.g., dimension and pose of the active objects), we can sort the predicted grasp pose of the objects and generate a collision-free trajectory to grasp and manipulate the objects, which will improve the grasp success rate further. The collision-avoidance topic is beyond the scope of this paper, and we refer the reader to our previous work on robot motion planning, in which we demonstrated how a robot could learn to avoid dynamic obstacles while performing a pick and place task~\cite{luo2021self}}. 

By comparing the results obtained from different rounds of experiments (isolated, packed, pile, and highly cluttered), we found that top-down grasps achieved mid-level success rate for isolated objects (see Table~\ref{table:isolated_exps}) but worked well for packed/piled objects (see Tables~\ref{packed_exps} and \ref{pile_exps}). The underlying reason was that in the case of the pile and packed scenarios, the center of packed items and also the center of the pile of objects was defined such that all objects can be reached from above, whereas in the case of the isolated object scenario, objects are randomly placed within the workspace. Consequently, in some cases the target object could not be reached from above. 
A video of these experiments is available online: \href{https://youtu.be/r7Ra8BJsAY4}{\cblue{{\texttt{https://youtu.be/r7Ra8BJsAY4}}}}


\section{Conclusion}

In this paper, we proposed a deep learning approach for real-time multi-view 3D object grasping in highly cluttered environments. We trained the approach in an end-to-end manner. The proposed approach allows robots to robustly interact with the environments in both isolated and highly crowded scenarios. In particular, for a given scene, our approach first generates three orthographic views, The best view is then selected and fed to the network to predict a pixel-wise grasp configuration for the given object. The robot is finally commanded to execute the highest-ranked grasp synthesis. To validate the performance of the proposed method, we performed extensive sets of real and simulation experiments in four everyday scenarios: \textit{isolated}, \textit{packed objects}, \textit{pile of objects}, and \textit{highly cluttered scenes}. Experimental results showed that the proposed method worked very well in all four scenarios, and outperformed the selected state-of-the-art approaches. In the continuation of this work, we would like to investigate the possibility of improving grasp performance by learning a shape completion function that receives a partial point cloud of a target object and generates a complete model. We then use the full model of the object to estimate grasp synthesis map. The other direction would be extending the proposed approach by adding an eye-in-hand system to have more flexibility to reconstruct parts of the environment that we are interested in. In particular, by moving the sensor to the desired areas, we can capture significantly more details that would otherwise not be visible. This information can be very helpful and lead to better grasp planning.

\bibliographystyle{unsrt}
\bibliography{literature}

\end{document}